\def\year{2020}\relax
\documentclass[letterpaper]{article} 
\usepackage{aaai20}  
\usepackage{times}  
\usepackage{helvet} 
\usepackage{courier}  
\usepackage[hyphens]{url}  
\usepackage{graphicx} 
\usepackage{graphicx}  
\newcommand{\citep}{\cite}
\newcommand{\citet}[1]
{\citeauthor{#1} \shortcite{#1}}

\usepackage{array,multirow}
\usepackage{enumitem}
\usepackage{graphicx}
\usepackage{subcaption}
\usepackage{amsmath}
\usepackage{amsfonts}
\usepackage{xcolor}
\usepackage{dsfont}
\usepackage{booktabs}
\usepackage{cancel}
\graphicspath{ {figs/} }

\usepackage{bm}

\def\dL2{{D_{\mathcal{L}_2}}}
\def\dCOS{{D_{\cos}}}
\def\dMMD{{D_{\operatorname{MMD}}}}
\def\dFLD{{D_{\operatorname{FLD}} }}
\def\dCORAL{{ D_{\operatorname{CORAL}} }}

\title{Multi-Source Domain Adaptation for Text Classification via DistanceNet-Bandits}

\author{Han Guo \;\;\;\;\;\;\; Ramakanth Pasunuru \;\;\;\;\;\;\; Mohit Bansal \\
  UNC Chapel Hill \\
  {\tt \{hanguo, ram, mbansal\}@cs.unc.edu} \\
 }

\begin{document}

\maketitle

\begin{abstract}
Domain adaptation performance of a learning algorithm on a target domain is a function of its source domain error and a divergence measure between the data distribution of these two domains. We present a study of various distance-based measures in the context of NLP tasks, that characterize the dissimilarity between domains based on sample estimates. We first conduct analysis experiments to show which of these distance measures can best differentiate samples from same versus different domains, and are correlated with empirical results. Next, we develop a DistanceNet model which uses these distance measures, or a mixture of these distance measures, as an additional loss function to be minimized jointly with the task's loss function, so as to achieve better unsupervised domain adaptation. Finally, we extend this model to a novel DistanceNet-Bandit model, which employs a multi-armed bandit controller to dynamically switch between multiple source domains and allow the model to learn an optimal trajectory and mixture of domains for transfer to the low-resource target domain. We conduct experiments on popular sentiment analysis datasets with several diverse domains and show that our DistanceNet model, as well as its dynamic bandit variant, can outperform competitive baselines in the context of unsupervised domain adaptation.
\end{abstract}

\section{Introduction}
\label{sec:intro}
In situations where large-scale annotated datasets are available, supervised learning algorithms have achieved remarkable progress in various NLP challenges~\cite{lecun2015deep}. Most supervised learning algorithms rely on the assumption that data distribution during training is the same as that during test. However, in many real-life scenarios, the data distribution of interest at test-time might be different from that during training. The process of collecting new datasets that reflect the new distribution is usually not scalable due to monetary as well as time constraints.
Hence, the goal of domain adaptation is to construct a learning algorithm, which, given samples of observations from a source domain, is able to adapt its performance to a target domain where the data distribution could be different.

Two major research areas in domain adaptation include supervised domain adaptation and unsupervised domain adaptation. In the former setup, limited training data from the target domain is available to provide supervision signals~\cite{daume2009frustratingly}, whereas in the latter case, only unlabeled data from the target domain is available~\cite{ganin2016domain,long2017deep,bousmalis2016domain,sun2016return,sun2016deep,tzeng2017adversarial}.
In this work, we focus on the unsupervised domain adaptation.
It has been shown that the domain adaptation performance is influenced by three major (and orthogonal) factors~\cite{ben2010theory}. The first factor is the model performance on the source task, which benefits from recent advancements in neural models and is orthogonal to our focus. 
The second factor is the difference in the labeling functions across domains, which is inherent to the nature of the dataset and expected to be small in practice~\cite{ben2010theory}. The third factor represents a measure of divergence of data distributions -- if the data distribution between the source and target domain is similar, we can reasonably expect a model trained on the source domain to perform well on the target domain.
Our work primarily focuses on the last factor and aims to study the following two questions in the context of NLP: how to accurately estimate the dissimilarity between a pair of domains (Sec.~\ref{sec:distances} and Sec.~\ref{sec:comparison-of-distances}), and how to leverage these domain dissimilarity measures to improve domain adaptation learning (Sec.~\ref{sec:models} and Sec.~\ref{sec:results}).

To this end, we first provide a detailed study (comparison, models, and analyses) of several domain distance measures from the literature, with the goal of scalability (easy to calculate), differentiability (can be minimized), and interpretability (in a simple analytical form with well-studied properties), namely $\mathcal{L}_2$,
Maximum Mean Discrepancy (MMD)~\cite{gretton2012kernel}, Fisher Linear Discriminant (FDA)~\cite{friedman2001elements}, Cosine, and Correlation Alignment (CORAL)~\cite{sun2016return}.
We start by defining these distance measures in Sec.~\ref{sec:distances}, and provide a set of analyses to assess them in Sec.~\ref{sec:comparison-of-distances}: (1) the ability of these distance measures to separate domains, and (2) the correlation between these distance measures and empirical results.
From these analyses, we note that there does not exist a single best distance measure that fits all, and each measure provides an estimate of domain distance that could be complementary (e.g., based on discrepancy versus class separation).
Thus, we also propose to use a mixture of distance measures, where we additionally introduce an unsupervised criterion to select the best distance measures so as to reduce the number of extra weight hyperparameters when mixing them.

Motivated by the aforementioned analysis, we next present a simple `DistanceNet' model (in Sec.~\ref{sec:models}) that integrates these measures into the training optimization. In particular, we augment the classification task loss function with an additional distance measure. By minimizing the representational distances between features from source and target domains, the model learns better domain-agnostic features.
Finally, when data from multiple source domains are present, we learn a dynamic scheduling of these domains that maximizes the learning performance on the no-training target task by framing the problem of dynamic domain selection as a multi-armed bandit problem, where each arm represents a candidate source domain.

We conduct our analyses and experiments on a popular sentiment analysis dataset with several diverse domains from~\citet{liu2017adversarial}, and present the domain adaptation results in Sec.~\ref{sec:results}. We first show
that a subset of the domain discrepancy measures is able to separate samples from source and target domains. Then we show that our DistanceNet model, which uses one or a mixture of multiple domain discrepancies as an extra loss term, can outperform multiple competitive baselines. Finally, we show that our dynamic, bandit variant of the DistanceNet can also outperform a fairly comparable multi-source baseline that has access to the same amount of data.

We start by reviewing related work in Sec.~\ref{sec:related-work}, and then introduce both the distance measures as well as the domain adaptation models in Sec.~\ref{sec:distances}-\ref{sec:models}. Finally, we present analyses on distance measures and experimental results in Sec.~\ref{sec:experimental-setup}-\ref{sec:results}.

\section{Related Work}
\label{sec:related-work}
Building an algorithm for domain adaptation is an open theoretical as well as practical problem~\cite{blitzer2006domain,pan2010survey,kulis2011you,glorot2011domain,blitzer2011domain,csurka2017domain,wang2018deep,chu2018survey,ponti2018isomorphic,murthy2018judicious,saito2018maximum,kuroki2019unsupervised,lee2019domain}.
When labeled data from target domain is available, supervised domain adaptation can achieve state-of-the-art results via fine-tuning, especially when source domain has orders of magnitude more data than target domain~\cite{zeiler2014visualizing,oquab2014learning,babenko2014neural,mccann2017learned,peters2018deep,devlin2018bert,radford2018improving,radford2019language}. For unsupervised domain adaptation (no labels for target domains), there exist multiple approaches that have achieved remarkable progress, such as instance selection/reweighting~\cite{huang2007correcting,gong2013connecting,remus2012domain} and feature space transformation~\cite{pan2011domain,baktashmotlagh2013unsupervised}. In this work we mainly focus on measuring domain discrepancy.

The works of~\citet{kifer2004detecting},~\citet{ben2007analysis}, and~\citet{ben2010theory} provide an upper bound on the performance of a classifier under domain shift.
They introduce the idea of training a binary classifier to distinguish samples from source/target domains, and the error $\mathcal{H}$-divergence provides an estimate of the discrepancy between domains. 
A tractable approximation, proxy $\mathcal{A}$-distance, applies a trained linear classifier to minimize a modified Huber loss~\cite{ben2007analysis}.

Recent works further aim to provide more efficient estimates of the domain discrepancy. One popular choice is matching the distribution means in the kernel-reproducing Hilbert space (RKHS)~\cite{borgwardt2006integrating,huang2007correcting,gong2013connecting,ghifary2014domain,tzeng2014deep,long2015learning,louizos2015variational,zhang2015deep,long2016unsupervised,bousmalis2016domain,zellinger2017central,long2017deep,yan2017mind,li2017demystifying,rozantsev2019beyond} using Maximum Mean Discrepancy (MMD)~\cite{gretton2012kernel}. These methods have also been used in generative models~\cite{li2015generative,dziugaite2015training}. Other methods explored in the literature include central moment discrepancy (CMD)~\cite{zellinger2017central}, correlation alignment (CORAL)~\cite{sun2016return,sun2016deep}, canonical correlation analysis (CCA)~\cite{blitzer2011domain}, cosine similarity~\cite{benaim2017one}, association loss~\cite{haeusser2017associative}, and metric learning~\cite{mahadevan2018unified}.
In addition to these directly-computable metrics, another successful approach is to encourage learned representations to fool a classifier~\cite{goodfellow2014generative} whose goal is to distinguish samples from the source domain and target domain~\cite{ganin2016domain,shen2017wasserstein}.

When multiple domain adaptation criteria are available,
~\citet{ruder2017learning} use Bayesian optimization to decide the choice of metric, and~\citet{ying2018transfer} use a meta-learning formulation. Also,~\citet{guo2018multi} and~\citet{chen2019multi} used a mixture-of-experts (MoE) approach for multi-source and multilingual domain adaptation.
In our work, we provide a comparison of multiple domain distance measures (introduced in statistical learning/vision communities) in the context of NLP classification tasks such as sentiment analysis, where we analyze the domain-separability skills of these metrics and then also explore multiple ways of integrating them into the training dynamics (e.g., using a mixture of these distance measures as an additional loss function and via a multi-armed bandit controller that can dynamically switch between several source domains during training, to allow the model to learn an optimal trajectory and mixture of domains).

Many problems can be cast as a multi-armed bandit problem. For example,~\citet{graves2017automated} use a multi-armed bandit (MAB)~\cite{bubeck2012regret} to learn a curriculum of tasks to maximize learning efficiency,~\citet{sharma2017online} use MAB to choose which domain of data to feed as input to a single model (in the context of Atari games), and~\citet{guo2019autosem} use MAB for task selection during multi-task learning of text classification. In our work, we instead use a MAB controller with upper confidence bound (UCB)~\cite{auer2002finite} for the task of multi-source domain selection for domain adaptation.


\section{Domain Distance Measures}
\label{sec:distances}
In Sec.~\ref{sec:intro}, we described that domain adaptation performance is related to domain distance/dissimilarity. Here, we will first describe our individual distance measures. Then we will describe our mixture of distances. Later in Sec.~\ref{sec:comparison-of-distances}, we will provide detailed analysis of these distance measures.
Given source domain samples $X_s = \{x^s_1, x^s_2, ..., x^s_{n_s}\}$ as well as target domain samples $X_t = \{x^t_1, x^t_2, ..., x^t_{n_t}\}$, where we assume $x \in \mathbb{R}^d$ are the embedding representations of the input data (e.g., sentences) produced from some feature extractors (e.g., LSTM-RNN), the goal of the distance measure is to estimate how different these two domains are. We will introduce five such methods: $\mathcal{L}_2$ distance, Cosine distance, Maximum Mean Discrepancy (MMD), Fisher Linear Discriminant (FLD), as well as CORAL.\footnote{We also experimented with proxy $\mathcal{A}$-distance from~\citet{ben2007analysis}, which scored favorably on most of our evaluations. However, due to its non-differential nature as well as high computation cost, we do not include it here.}

\subsection{$\mathcal{L}_2$ Distance}

The $\mathcal{L}_2$ distance measures the Euclidean distance between source domain and target domain samples. Define $\mu_s = \frac{1}{n_s}\sum_i x^s_i$ and $\mu_t = \frac{1}{n_t}\sum_i x^t_i$, the $\mathcal{L}_2$ distance is: $\dL2 (X_s, X_t) = \| \mu_s -  \mu_t \|_2$.

\subsection{Cosine Distance}
Cosine similarity is a measure of similarity between two vectors of an inner product space that measures the cosine of the angle of these vectors: $S_{\cos} = \frac{\mu_s \cdot \mu_t}{\|\mu_s\|_2 \|\mu_t\|_2}$, and cosine distance is $\dCOS = 1 - S_{\cos}$.

\subsection{Maximum Mean Discrepancy (MMD)}
Given two sets of source domain and target domain samples independently and identically distributed (i.i.d.) from $P_s(X)$ and $P_t(X)$, respectively. The statistical hypothesis testing is used to distinguish between the null hypothesis $\mathcal{H}_0 {:} P_s {=} P_t$, and the alternative hypothesis $\mathcal{H}_A {:} P_s {\ne} P_t$ via comparing test statistic, which is described next.
Maximum Mean Discrepancy or MMD~\cite{gretton2012kernel}, also known as kernel two-sample test, is a frequentist estimator for answering the above question. MMD works by comparing statistics between the two samples, and if they are similar then they are likely to come from the same distribution. This is known as an integral probability metric (IPC)~\cite{muller1997integral} in statistics literature. Formally, let $\mathcal{F}$ be a class of functions $f : \mathcal{X} \rightarrow \mathbb{R}$, and the maximum mean discrepancy is:
\begin{equation*}
\operatorname{MMD}_{\mathcal{F}}[P_s, P_t] = \sup_{f \in \mathcal{F}} \mathbb{E}_{x^s}[ f(x^s)] - \mathbb{E}_{x^t}[f(x^t)]
\end{equation*}
Note that this equation involves a maximization over a family of functions. However,~\citet{gretton2012kernel} show that when the function class $\mathcal{F}$ is the unit ball in a reproducing kernel Hilbert space (RKHS) endowed with a characteristic kernel $k$, this can be solved in closed form. A corresponding unbiased finite sample estimate is:
\begin{equation*}
    \begin{split}
    & \operatorname{MMD}^2_{\mathcal{F}}[P_s, P_t]
    =\frac{1}{n_s^{2}} \sum_{i=1}^{n_s} \sum_{i^{\prime}=1}^{n_s} k\left(x^s_{i}, x^s_{i^{\prime}}\right) \\
    &\mbox{-}\frac{2}{n_s n_t} \sum_{i=1}^{n_s} \sum_{j=1}^{n_t} k\left(x^s_{i}, x^t_{j}\right)
    \mbox{+}\frac{1}{n_t^{2}} \sum_{j=1}^{n_t} \sum_{j^{\prime}=1}^{n_t} k\left(x^t_{j}, x^t_{j^{\prime}}\right)
    \end{split}
\end{equation*}
For universal kernels like the Gaussian kernel $k\left(x, x^{\prime}\right)=\exp \left(-\frac{1}{2 \sigma}\left|x-x^{\prime}\right|^{2}\right)$ with bandwidth $\sigma$, minimizing MMD is analogous to minimizing a distance between all moments of the two distributions~\cite{li2015generative}. Here we will use $\dMMD (X_s, X_t) = \operatorname{MMD}^2_{\mathcal{F}}[P_s, P_t]$.

\subsection{Fisher Linear Discriminant}

Fisher linear discriminant analysis (FLD)~\cite{friedman2001elements} finds a projection (parameterized by $w$) where class separation is maximized. In particular, the goal of FLD is to give a large separation of class means while simultaneously keeping in-class variance small. This is formulated as $w^\star = \arg\max_w J(w) = \arg\max_w \frac{w^\mathrm{T} S_B w}{w^\mathrm{T} S_W w}$, where $S_B$ is the between-class covariance matrix which is defined as $S_B = (\mu_s - \mu_t) (\mu_s - \mu_t)^T$, $S_W$ is the within-class covariance matrix which is defined as $S_W = \sum_{c \in \{0, 1\}} \sum_i (x^{(c)}_i - \mu_c) (x^{(c)}_i - \mu_c)^\mathrm{T}$, $\mu_c$ is the class mean and $\{0, 1\}$ here refers to source/target domain. The optimal $w^\star$ can be solved analytically as: $w^\star \propto S_W^{-1}(\mu_1 - \mu_2)$.
Though the optimal $w^\star$ is usually desired, here we use the optimal $J$ as a proxy of domain distance, and thus define our Fisher distance as $\dFLD (X_s, X_t) {=} J(w^\star)$, which is a measure of difference between source/target representation means normalized by a measure of within-class scatter matrix. Note that computing the $\dFLD$ is analogous to approximating the divergence between two domains by training an FLD to discriminate between unlabeled instances from source and target domains.

\subsection{Correlation Alignment (CORAL)}
The CORAL (correlation alignment)~\cite{sun2016deep,sun2016return} loss is defined as the distance between the second-order statistics of the source and target samples: $\dCORAL (X_s, X_t) {=} \frac{1}{4 d^{2}}\left\|C_{s}-C_{t}\right\|_{F}^{2}$, where $\|\cdot\|^2_{F}$ denotes the squared matrix Frobenius norm, $d$ represents feature dimension, and $C_s$ and $C_t$ are the covariance matrices of source and target samples.

\subsection{Mixture of Distances}
\label{subsec:mixture-of-distances}
As we will demonstrate in Sec.~\ref{sec:comparison-of-distances}, no single distance measure outperforms all the others in our analyses. Also, note that while different distance measures provide different estimates of domain distances, each distance measure has its pathological cases. For example, samples from a Gaussian distribution and a Laplace distribution with same mean and variance might have small $\mathcal{L}_2$ distances even though they are different, whereas MMD can differentiate between them~\cite{gretton2012kernel}. It is thus useful to consider a mixture of distances:
\begin{equation}
    D_{m} (X_s, X_t) = \sum_k \alpha_k D_k (X_s, X_t)
\label{eq:mixture-of-distances}
\end{equation}
where $\alpha_k \in \mathbb{R}$ is the coefficient for $k$-th distance. While appealing at first, naively adding all the distance measures to the mixture introduces unnecessary hyper-parameters. In Sec.~\ref{subsec:importance-of-measure-components}, we will introduce simple unsupervised criteria to only include a subset of these distance measures.

\section{Models}
\label{sec:models}

\begin{figure}
\centering
\includegraphics[width=0.95\linewidth]{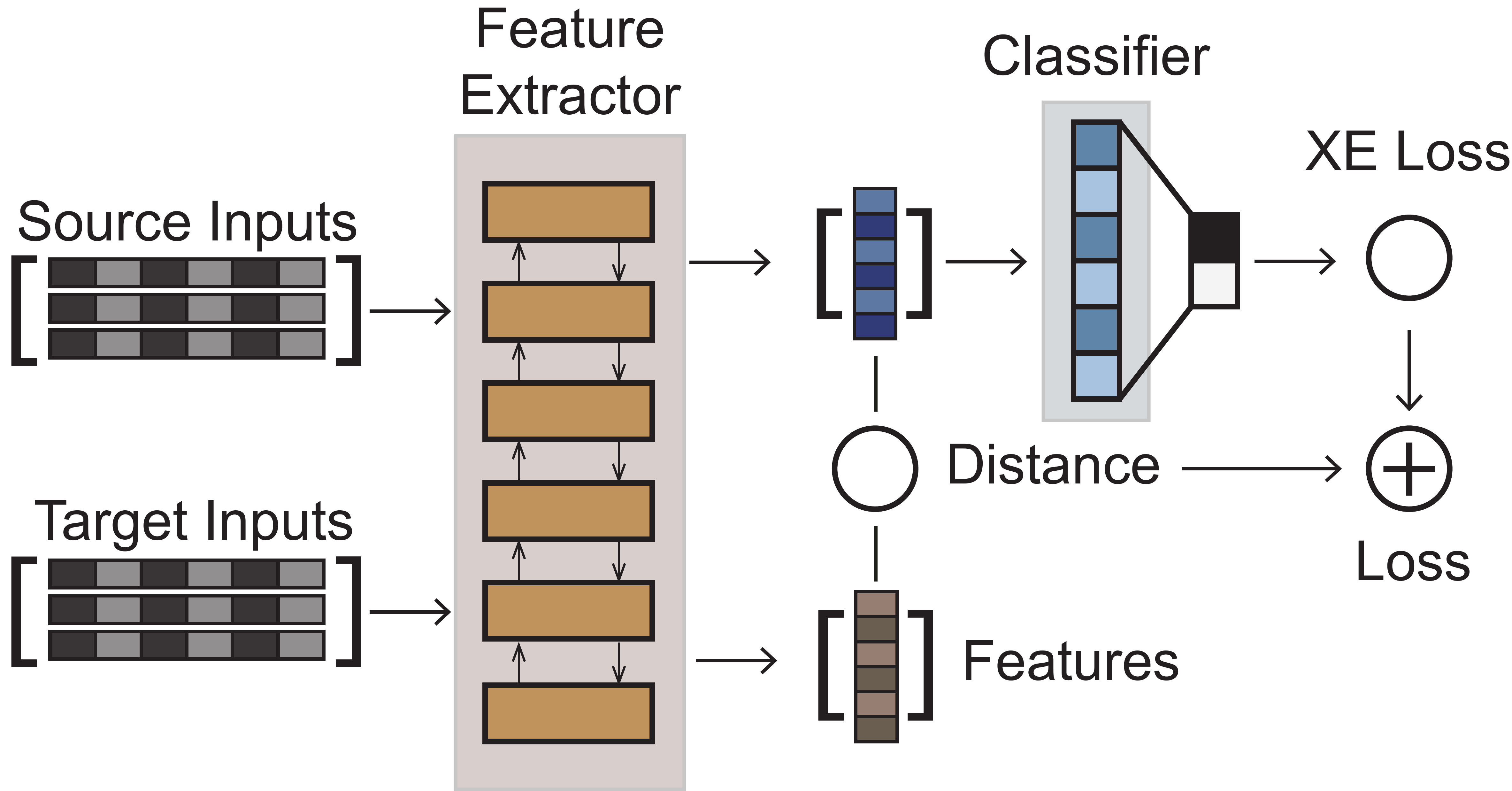}
\caption{Overview of the DistanceNet model. The model takes both the source domain data (labeled) and target domain data (unlabeled), and computes feature representations. The distances, calculated via the distance measures, between the source and target samples are added to the (source) cross-entropy loss to be minimized jointly during the training.
\label{fig:baseline}
}
\end{figure}

We will first describe the baseline and our DistanceNet model (based on a single source domain) which actively minimizes the distance between the source and target domain during the model training for domain adaptation. Then we introduce the multi-source variant of DistanceNet that additionally utilizes a multi-armed bandit controller to learn a dynamic curriculum of multiple source domains for training a domain adaptation model.

\subsection{Baseline Model}
\label{subsec:baseline-model}
Given a sequence of tokens $\{w_0, w_1, ..., w_T\}$, we first embed these tokens into vector representations $\{e_0, e_1, ..., e_T\}$. Let $h_T = \operatorname{LSTM}(\{e_t\}, \theta_1)$ be the output of the LSTM-RNN parameterized by $\theta_1$. The probability distribution of labels is produced by $\hat{y} = \operatorname{FC}(h_T, \theta_2)$, where $\operatorname{FC}$ is a fully connected neural network with parameters $\theta_2$. The model is trained to minimize the cross entropy between predicted outputs $\hat{y}$ and ground truth $y$ with $N$ training examples and $C$ classes: $L_{\text{XE}}(\hat{y}, y)=-\sum_{i=1}^{N} \sum_{j=1}^{C} y_{i,j} \log \hat{y}_{i,j}$.

\subsection{DistanceNet}
The work of~\citet{ben2010theory} shows that domain adaptation performance is related to source domain performance and source/target domain distance. The first part (source domain performance) is already handled by the cross entropy loss (Sec.~\ref{subsec:baseline-model}), and it is thus natural to additionally encourage the model to minimize the representational distances between source and target samples. To that end, we augment the classification task's loss function with a domain distance term.
Given a sequence of tokens from the source domain $\{w^s_0, w^s_1, ..., w^s_{T_s}\}$, a sequence of tokens from the target domain $\{w^t_0, w^t_1, ..., w^t_{T_t}\}$, and model parameterized by $(\theta_1, \theta_2)$, the new loss function for our DistanceNet (see Fig.~\ref{fig:baseline}) is then:
\begin{equation}
    L(\hat{y}^s, y^s) = L_{\text{XE}}(\hat{y}^s, y^s) + \beta D_k(h^s_{T_s}, h^t_{T_t})
\label{eq:loss}
\end{equation}
where $\hat{y}^s, y^s$ are the predicted and ground truth outputs of source domain,  $h^s_{T_s}, h^t_{T_t}$ are the representations of source and target domain, and $D_k$ is the choice of distance measure from Sec.~\ref{sec:distances}.

\subsection{Dynamic Multi-Source DistanceNet using Multi-Armed Bandit}

\begin{figure}
\centering
\includegraphics[width=0.95\linewidth]{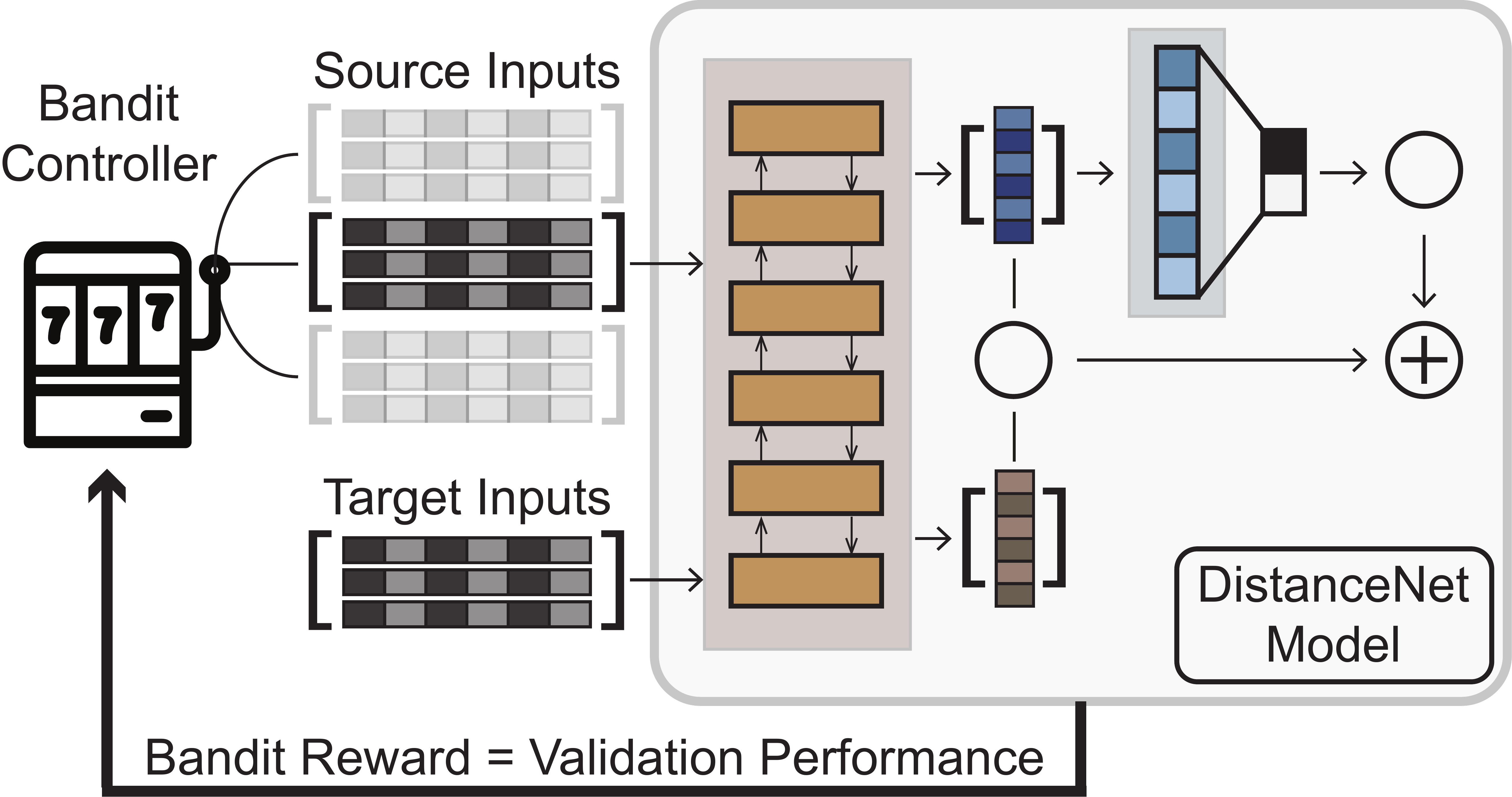}
\caption{Overview of our multi-source DistanceNet model with controller. During the training, a multi-armed bandit controller dynamically selects the source domain from a set of candidate source domains. The controller updates its belief over the utility of each domain via receiving feedback on validation set.
\label{fig:bandit}
}
\end{figure}

In the previous section, we described our method for fitting a model on a pair of source/target domains. However, when we have access to multiple source domains, we need a better way to take advantage of these extra learning signals. One simple method is to treat these multiple source domains as a single (big) source domain, and apply algorithms described previously as usual. 
But as the model representation changes throughout the training, the domain that can provide the most informative training signal might change over time and based on the training curriculum history. This is also related to learning importance weights~\cite{ben2010theory} of each source domain over time for the target domain.
Thus, it might be more favorable to dynamically select the sequence of source domains to deliver the best outcome on the target domain task.

Here, we introduce a novel multi-armed bandit controller for dynamically changing the source domain during training (Fig.~\ref{fig:bandit}).
We model the controller as an $M$-armed bandits (where $M$ is the number of candidate domains) whose goal is to select a sequence of actions/arms to maximize the expected future payoffs. At each round, the controller selects an action (candidate domain) based on noisy value estimates and observes a reward. More specifically, 
as the training progresses, the controller picks one of the training domains and have the task model train on the selected domain
using the loss function specified in Eq.~\ref{eq:loss}, and the performance on the validation data will be used as the reward provided to the bandit as feedback. We use upper confidence bound (UCB)~\cite{auer2002finite} bandit algorithm, which chooses the action (i.e., the source domain to use next) based on the performance upper bound: $a_{t}^\text{UCB}=\arg \max _{a \in \mathcal{A}} Q(a)+\sqrt{\frac{2 \log t}{N_{t}(a)}}$, where $a_t$ represents the action at iteration time $t$, $N_{t}(a)$ counts the number of times the action has been selected, and $\mathcal{A}$ represents the set of candidate actions (i.e., the set of candidate source domains). $Q(a)$ represents the action-value of the action, and is calculated as the running average of rewards.\footnote{
One could also consider weighting each domain based on the distances, but these keep changing as DistanceNet's training evolves (which minimizes the distance). Further, our bandit decides the arm to pull based on DistanceNet's performance, thus already behaving similar to the distance-weighting approach (while also automatically learning these weights as a curriculum).}


\begin{figure*}[t]
\centering
\includegraphics[width=0.95\linewidth]{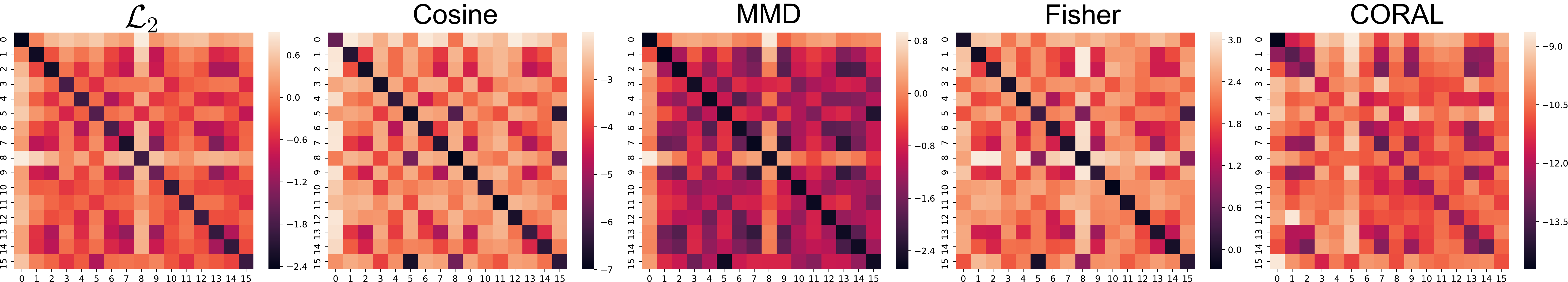}
\caption{Domain separability test of distance methods. The order is (left to right): $\dL2, \dCOS, \dMMD, \dFLD, \dCORAL$. The value of each entry at position $(i, j)$ refers to the distance between samples from $i$-th source domain and $j$-th target domain. In particular, the entries on the diagonal refer to the in-domain distances, and off-diagonal entries refer to the between-domain distances. Values shown are the $\log$ of the distances for visualization purposes. 
\label{fig:distances}
}
\end{figure*}

\section{Experimental Setup}
\label{sec:experimental-setup}

\noindent\textbf{Dataset:} We evaluate our methods using the datasets collected by~\citet{liu2017adversarial}\footnote{The datasets include ``unlabeled'' split.}, which contains $16$ datasets of product reviews~\cite{blitzer2007biographies} and movie reviews~\cite{maas2011learning,pang2005seeing}, where the task is to classify these reviews as positive or negative. The performance of a model on this task is measured by accuracy.
Since the number of experiments scales $\mathcal{O}(n^2)$ and $\mathcal{O}(n)$ for single- and multi-source experiments, we only evaluate on $3$ and $5$ datasets\footnote{MR, Apparel, Baby for single-source experiments. MR, Apparel, Baby, Books, Camera for multi-source experiments.} for experiments in Sec.~\ref{sec:results}, respectively.\footnote{Note that for $n$ tasks, there will be $n{\times}(n-1)$ source/target domain pairs experiments, and $n$ multi-source/single-target domain pairs experiments.} 
However, we still use the full set of domains for the analysis in Sec.~\ref{sec:comparison-of-distances}.

\noindent\textbf{Training Details:} Our baseline model is similar to that of~\citet{liu2017adversarial}. We use a single-layer Bi-directional LSTM-RNN as sentence encoder and a two-layer fully-connected with $\operatorname{ReLU}$ non-linearity layer to produce the final model outputs. The word embeddings are initialized with GloVe~\cite{pennington2014glove}. We train the model using Adam optimizer~\cite{kingma2014adam}. Following~\citet{ruder2017learning} and~\citet{bousmalis2016domain}, we chose to use a small number of target domain examples as validation set (for both tuning as well as providing rewards for the multi-armed bandit controller).\footnote{Note that the two models in Table~\ref{table:multi-source-distance-net} should be fairly comparable, since they have access to the same validation dataset for tuning or ``refining'' their hyper-parameters or as weak reward feedback. Further, there are scenarios in which querying the scalar rewards on a small validation dataset is easier than accessing the rich gradient information through them~\cite{bousmalis2016domain}. 
}
We use the adaptive experimentation platform Ax\footnote{\url{https://github.com/facebook/Ax}} to tune the rest of the hyperparameters and the search space for these hyperparameters are: learning rate $\in (10^{-4}, 10^{-3})$, dropout rate $\in (0.25, 0.75)$, $\beta \in (0.01, 1.0)$, and $\alpha_k \in (0.0, 1.0)$. We run each model for $3$ times. We use the average validation performance as our validation criteria, and report average test performance.


\section{Analysis of Distance Measures}
\label{sec:comparison-of-distances}
Given our 5 distance measures (described in Sec.~\ref{sec:distances}), we first want to ask which of these distance measures are able to measure domain (dis)similarities. 
Specifically, we conduct experiments to answer the following questions:\\
Q1. Is the distance measure able to differentiate samples from the same versus different domains?\\
Q2. Does the distance measure correlate well with empirical results?

These two questions are answered next in Sec.~\ref{subset:two-sample-test} and Sec.~\ref{subsec:corr-with-results}, respectively. After that, we will describe our unsupervised criteria for choosing a subset of distance measures (Sec.~\ref{subsec:importance-of-measure-components}) to be used in the mixture of distance measures introduced in Sec.~\ref{subsec:mixture-of-distances}.

\subsection{Domain Separability Test}
\label{subset:two-sample-test}
Given two sets of source and target domain samples: $X_s{=}\{x^s_1, x^s_2, ..., x^s_{n_s}\}$ and $X_t{=}\{x^t_1, x^t_2, ..., x^t_{n_t}\}$, which are independently and identically distributed (i.i.d.) from $P_s(X)$ and $P_t(X)$, respectively. The goal here is to find whether these samples come from the same domain or not.
For this, we compute the distance between the source and target samples, $d_k(P_s, P_t)$, via distance measure $D_k$ (selected from the distance measures defined in Sec.~\ref{sec:distances}):
\begin{equation}
    d_k (P_s, P_t) = \mathop{\mathbb{E}}_{x_s \sim P_s, x_t \sim P_t}[D_k(x_s, x_t)]
\end{equation}
For distance measure to estimate domain similarity, we expect $d_k (P_s, P_t)$ to be low when $P_s{=}P_t$, and high otherwise (similar to two sample test statistic~\cite{gretton2012kernel}).

Fig.~\ref{fig:distances} visualizes the results of our experiments, where the distance between exhaustive source/target domain pairs are measured on $16$ datasets. We take $200$ examples from each domain\footnote{We take source domain samples from the training set and target domain samples from the validation set to avoid overlapping examples when sampling from the same domain.}, and embed the sentences using pre-trained model\footnote{\url{https://tfhub.dev/google/tf2-preview/nnlm-en-dim128/1}}, after which the distances are calculated. 
In particular, the entries on the diagonal refer to the in-domain distances (i.e., source and target domain is the same), and off-diagonal entries refer to the between-domain distances. As we want the in-domain distances to be small and between-domain distances to be large, we expect the visualization of a good distance measure to have a dark line on the diagonal (indicating low values) and bright otherwise. From the visualization plots (Fig.~\ref{fig:distances}), we can see that $\dL2, \dCOS, \dMMD$ and $\dFLD$, are able to separate domains well. However, all these measures have different scales and sensitivity, hence, we next define two statistics to quantitatively compare different distance measures $d_k$, which are denoted by $z_1$ and $z_2$ corresponding to method-1 and method-2, respectively. These statistics are shown in Table.~\ref{table:distance}. We can see that most of these methods are able to separate domains, with the exception of $\dCORAL$. Next, we describe these methods.

\begin{table}
\centering
\small
\begin{tabular}{lccc}
\toprule
Name     & Method-1 & Method-2 & Result-Corr \\
\midrule
$\mathcal{L}_2$ &  1.00 &  6.35 $\times 10^{-3}$ & 0.67 \\
 Cosine         &  0.94 &  7.49 $\times 10^{-3}$ & 0.79 \\
    MMD         &  0.94 &  6.83 $\times 10^{-3}$ & 0.59 \\
 Fisher         &  0.88 &  5.99 $\times 10^{-3}$ & 0.65 \\
  CORAL         &  0.75 &  9.11 $\times 10^{-3}$ & 0.39 \\
\bottomrule
\end{tabular}
\caption{Distance comparison statistics and result-correlations. Note that for Method-1 and Result-Corr, higher numerical values are better, however, for Method-2, lower numerical values are better.
}
\label{table:distance}
\end{table}

\paragraph{Method-1.} The first method assess whether distances between samples from the same domain are lower than those between the samples from different domains, $d_k(P_i, P_i) \leq d_k(P_i, P_j) \forall i \ne j$. This statistic is appealing because it is invariant to scaling and translation, but does not concern how smaller in-domain distances are w.r.t. off-domain distances. Specifically, we compute the $z_1$ as:
\begin{equation*}
    \begin{split}
    z_1(d_k) = \frac{1}{N} \sum_i \mathbb{I} \big[& d_k(P_i, P_i) \leq d_k(P_i, P_j) \land \\
                                             & d_k(P_i, P_i) \leq d_k(P_j, P_i) \forall i \ne j \big]
    \end{split}
\end{equation*}
We can see that $\dL2$ achieves the highest score, whereas $\dCORAL$ achieves the lowest.

\paragraph{Method-2} The second method assesses how smaller the value of $d_k(P_i, P_i)$ is in comparison to $ d_k(P_i, P_j) \forall i \ne j$. To compute this, we first standardize\footnote{Subtracting the means and normalize by standard deviations.} the matrix $\{d_k(P_i, P_j), \forall i,j \}$, and then apply $\operatorname{softmax}$ function to ensure that all entries are positive. Then we compute the sum of the diagonal entries of the transformed matrix $\{d_k^\prime(P_i, P_j), \forall i,j \}$ as our second quantitative assessment ($z_2$, note that smaller is better):
\begin{equation}
    z_2(d_k) = \sum_i d_k^\prime(P_i, P_i)
\end{equation}
We can see that $\dFLD$ obtains the lowest/best value, and $\dCORAL$ scores the largest value.

\subsection{Correlation With Results}
\label{subsec:corr-with-results}

The methods described previously answer the question of whether $P_s{=}P_t$ given the samples $X_s$ and $X_t$. However, the assessment we are interested in ultimately is whether the distance measures correlate with the true domain distances. As the true domain distance is latent, here we will use a proxy. We denote $r(P_s, P_t)$ as the performance of the baseline model trained on the source domain and evaluated on the target domain. We want to measure the correlation between $d_k(P_s, P_t)$ and $r(P_s, P_t)$. Specifically, we train and evaluate baseline models on all source/target domain pairs, and then compute the Pearson correlation coefficient between the results (averaged over three runs) and distance measures. 
The values are shown in Table~\ref{table:distance}, where we can see that most of the distance measures are correlated with actual performance, with $\dCORAL$ having the lowest correlation with empirical performance (hence we ignore $\dCORAL$ for all future experiments, given that it is the worst by large margins on all 3 analysis methods above).

\subsection{Informativeness of Mixture Components}
\label{subsec:importance-of-measure-components}
Lastly, we present the basis for deciding which distances to (not) include in the mixture formulation described in Sec.~\ref{subsec:mixture-of-distances}. 
Specifically, our goal is to remove redundant distance measures from the mixture, subject to the constraint that the reduced mixture still provides sufficient information about the distances between two domains. We approach this problem via estimating the `informativeness' of each distance measure. This is analogous to influence functions, a classic technique from robust statistics~\cite{cook1980characterizations,koh2017understanding}.
To motivate our approach, let's say our mixture $\{D_k\}^K_{k=1}$ includes all distance measures which are previously defined (Sec,~\ref{sec:distances}). Suppose we have a function $\phi(\{D_k\}^K_{i=1})$ which can give us an estimate of the quality of the mixture. Now, we proceed by removing one metric (say $D_m$) from the mixture and apply the function $\phi$ to give us an estimate of the quality of the reduced mixture, $\phi(\{D_k\}^K_{k\ne m})$. We can now define an estimate of distance measure's informativeness:
\begin{equation}
    \mathcal{I}(D_m) = \phi(\{D_k\}^K_{i=1}) - \phi(\{D_k\}^K_{i\ne m})
\label{eq:importance}
\end{equation}
If $\mathcal{I}(D_m)$ is small, we can say the removed metric is not informative given other components in the mixture. 
Here, we use the optimal $z_2$ statistics\footnote{We do not use $z_1$ because it is not differentiable (calculated as multiple binary comparisons), and $z_1$ already achieved almost-maximum scores (Table~\ref{table:distance}) thus making the optimization less useful. Also, since we evaluate using an unsupervised criterion, we decided not to use correlation because it is a supervised evaluation.} (which is unsupervised) defined in Sec.~\ref{subset:two-sample-test} as the mixture evaluation function:
\begin{equation*}
    \phi(\{D_k\}^K_{i=1}) = \max_{\alpha_1, ..., \alpha_k} z_2\Big(\sum_k \alpha_k D_k (X_s, X_t)\Big)
\end{equation*}
where we estimate the maximum value using gradient descent (via the JAX library). 
We found that removing $\dCOS$ has far lower impact on the optimal $z_2$, and thus in our experiments using mixture of distances, we do not include $\dCOS$ (see Table~\ref{table:empitical-importance} for detailed scores of informativeness for all the distance measures).

\begin{table}
\centering
\small
\begin{tabular}{lc}
\toprule
Name     & Informativeness Estimate \\
\midrule
$\mathcal{L}_2$ &  -2.086 $\times 10^{-3}$ \\
 Cosine         &  -0.001 $\times 10^{-3}$ \\
    MMD         &  -1.775 $\times 10^{-3}$ \\
 Fisher         &  -0.024 $\times 10^{-3}$ \\
\bottomrule
\end{tabular}
\caption{Estimated importance of each distance measure as a component in the mixture. Values are negative because the full mixture achieves the lowest $z_2$. Lower (more negative) value means the distance measure provides more information. We did not include $\dCORAL$ because it scored unfavorably in our previous assessments.}
\label{table:empitical-importance}
\end{table}


\section{DistanceNet and Bandit Results}
\label{sec:results}
In this section, we show domain-adaptation experimental results for the sentiment classification task on the target domain (using out-of-domain source training data). We start with comparing our (in-domain) baseline to previous work, where the source and target domain are the same. Then we will show the results of our DistanceNet (with both single distance and mixture-of-distance measures), when the source domain and target domain is different. Lastly, we will show the results of our multi-source DistanceNet baseline versus our multi-source DistanceNet bandit model which dynamically selects source domains.
Based on the results of Sec.~\ref{sec:comparison-of-distances}, we do not include $\dCORAL$ in our DistanceNet experiments, and do not include both $\dCORAL$ and $\dCOS$ in our DistanceNet with mixture-of-distance experiments.

\begin{table}
\centering
\small
\begin{tabular}{lcccccc}
\toprule
Model                      & MR   & Aprl & Baby & Books & Camera \\
\midrule
Liu \shortcite{liu2017adversarial} & 74.7 & 86.0    & 83.5 & 81.0 & 86.0 \\
Ours                               & 73.8 & 87.2    & 85.2 & 81.4 & 88.1 \\
\bottomrule
\end{tabular}
\caption{Performance of our baseline compared with previous work~\cite{liu2017adversarial}.
}
\label{table:baseline}
\end{table}

\begin{table}
\centering
\small
\begin{tabular}{ccccccc|c}
\toprule
Source & \multicolumn{2}{c}{MR(M)} & \multicolumn{2}{c}{Aprl(A)} & \multicolumn{2}{c|}{Baby(B)} & Avg \\
\midrule
Target &
    \multicolumn{1}{c}{A} & \multicolumn{1}{c|}{B} &
    \multicolumn{1}{c}{M} & \multicolumn{1}{c|}{B} &
    \multicolumn{1}{c}{M} & \multicolumn{1}{c|}{A} & {} \\
\midrule

{\scriptsize DataSel}        &                    68.1 &  65.2 &    64.3 &  74.3 &  65.6 &    78.9 & 69.39 \\
{\scriptsize DANN}           &                    69.9 &  65.3 &    63.7 &  78.2 &  65.5 &    80.0 & 70.46 \\
\midrule
{\scriptsize Baseline}       &                    67.3 &  66.5 &    65.8 &  78.2 &  64.6 &    78.1 & 70.08 \\
{\scriptsize $\mathcal{L}_2$}&                    70.9 &  66.5 &    64.7 &  76.6 &  65.3 &    78.2 & 70.37 \\
{\scriptsize Cosine}         &                    70.2 &  66.2 &    64.6 &  78.3 &  65.3 &    78.2 & 70.48 \\
{\scriptsize MMD}            &                    69.9 &  67.1 &    64.3 &  77.1 &  66.0 &    78.1 & 70.42 \\
{\scriptsize Fisher}         &                    69.1 &  64.2 &    64.6 &  77.9 &  65.4 &    79.4 & 70.10 \\
{\scriptsize Mixture}        &                    70.4 &  67.1 &    65.6 &  79.0 &  66.5 &    79.3 & 71.32 \\
\bottomrule
\end{tabular}
\caption{Performance comparison of previous works (DataSel:~\citet{remus2012domain} 
; DANN:~\citet{ganin2016domain}), single-source baseline, and DistanceNet models.  }
\label{table:distance-net}
\end{table}

\begin{figure*}[t]
\centering
\includegraphics[width=\linewidth]{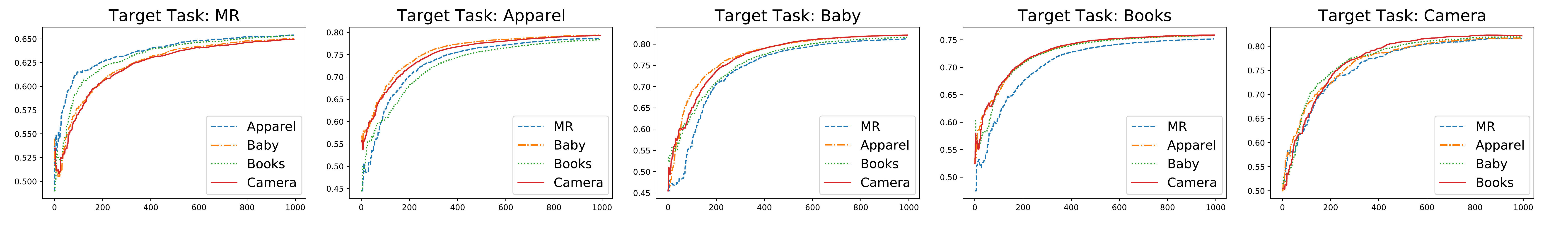}
\caption{Visualization examples of multi-armed bandits. Each line represents an arm (a source domain), X-axis refers to time, and Y-axis refers to the values of each arm (higher value of an arm corresponds to potentially more usefulness of the task). 
\label{fig:bandit-visualization}
}
\end{figure*}

\subsection{Single Source DistanceNet Results}

\paragraph{Baseline Results.}
In Table~\ref{table:baseline} we show the results of our (in-domain) baseline compared with similar models in~\citet{liu2017adversarial}. We can see that our baseline is stronger than comparable previous work in four of the five domains we considered.

\paragraph{DistanceNet Results.}
Table~\ref{table:distance-net} shows the results of baselines and DistanceNet models when the source and target domain is different, where the last column shows the average results.\footnote{Note that the single-distance methods, e.g., MMD, have been used in previous works~\cite{bousmalis2016domain,tzeng2014deep,benaim2017one} and can also be considered as baselines.}
First, comparing the numbers to those in Table~\ref{table:baseline}, we can see that performance drops when there is a shift in the data distribution. Next, we can see that by adding our domain distance measure as an additional loss term, the model is able to reduce the gap between in-domain performance and out-of-domain performance. In particular, all of our models perform better than our baseline in terms of average results, with MMD model better than the baseline by one corresponding standard deviation.\footnote{To calculate the standard deviation of the average results, we first compute the average results for each run, and compute the standard deviation of the average results. This is equivalent to computing the standard deviation of a single large prediction by concatenating model outputs for all tasks as a single output.}

\paragraph{Mixture DistanceNet Results.}
Table~\ref{table:distance-net} shows the results of our DistanceNet with mixture of distance measures experiments. From the results, we can see that leveraging the power of multiple distance measures additionally improves the results in out-of-domain settings, and achieving the highest average results (better than baseline by two standard deviations). We also compare our DistanceNet models to other domain-adaptation approaches. DANN encourages similar latent features by augmenting the model with a few standard layers and a new gradient reversal layer~\cite{ganin2016domain}. DataSel instead relies on data selection
based on domain similarity and complexity variance~\cite{remus2012domain}. From the results, we can see that our DistanceNet with mixture of distance measures outperforms these approaches (better w.r.t. standard deviation margins).

\subsection{Multi-Source DistanceNet-Bandit Results}
Table~\ref{table:multi-source-distance-net} shows the results for our multi-source experiments, where the source domains include all but the target domain, thus we have one result for each target domain. 
Here the baseline is the DistanceNet with mixture of distance measures, which selects domains in a round-robin fashion. Our model instead applies a dynamic controller to select the source domain to use. 
We can see from the results that using the dynamic controller improves the individual results, and the average results (better by two standard deviations).\footnote{Our single-source experiments suggested that ``MR'' and ``Books'' are not helpful for the learning of the other three tasks, thus we mask the DistanceNet loss from these domains when the target domain is not ``MR'' or ``Books''.}
In general, we observed that the bandit always improves over the non-bandit baseline (with two std. deviations) even when we simply reuse the best hyperparameters found in the single-source experiments, and when we employ a bandit without the DistanceNet loss (i.e., just cross-entropy).

\subsection{Multi-Armed Bandit Visualization}

Fig.~\ref{fig:bandit-visualization} provides example visualizations of the usefulness of each source domain for a given target domain during the training trajectory of multi-source bandit experiments. We provide a brief summary of our observations from these examples here. When the target task is ``MR'', we observed that ``Books'' and ``Apparel'' are more beneficial. When the target task is ``Apparel'', we found that ``Camera'' as well as ``Baby'' are beneficial; moreover, there the bandit learns to switch between ``Books'' and ``MR'' over time. 
When the target task is ``Baby'', we see that ``Camera'' and ``Apparel'' are beneficial.
When ``Books'' is the target task, we found that ``MR'' seemed to be less helpful. Finally, when the target-task is ``Camera'',  we see that ``Books'' had the highest value.

\begin{table}
\centering
\small
\begin{tabular}{lccccc|c}
\toprule
Model       & MR   & Aprl & Baby & Books & Camera & Avg \\
\midrule
Mixture    & 69.8 & 80.8 & 82.5 & 77.0 & 80.9 & 78.20 \\
+Bandit       & 72.0 & 82.3 & 82.8 & 78.3 & 81.3 & 79.30 \\
\bottomrule
\end{tabular}
\caption{Multi-source DistanceNet versus bandit. 
}
\label{table:multi-source-distance-net}
\end{table}


\section{Conclusion}
In this work, we presented a study of multiple domain distance measures to address the problem of domain adaptation. We provided analyses of these measures based on their ability to separate same/different domains and correlation with results. Next, we introduced our model, DistanceNet, which augments the loss function with the distance measures. Later, we extended our DistanceNet to the multi-source setup via a multi-armed bandit controller. Our experiment results suggest that our DistanceNet, as well as its variant with the multi-armed bandit, is able to outperform corresponding baselines.

\section*{Acknowledgments}
We thank the reviewers and Boyang Li for their helpful comments. This work was supported by DARPA (YFA17-D17AP00022),
NSF-CAREER Award \#1846185, ONR Grant \#N00014-18-1-2871, Google, Facebook, Baidu, and Salesforce.  
The views contained in this article are those of the authors and not of the funding agency.

\bibliography{main.bib}
\bibliographystyle{aaai}

\end{document}